# What is Hiding in Medicine's Dark Matter? Learning with Missing Data in Medical Practices

Neslihan Suzen
*School of Computing and Mathematical Sciences*
*University of Leicester*
Leicester, UK
ns553@leicester.ac.uk

Evgeny M. Mirkes
*School of Computing and Mathematical Sciences*
*University of Leicester*
Leicester, UK
em322@leicester.ac.uk

Damian Roland
*Children's Emergency Department, Leicester Royal Infirmary*
*University of Leicester*
Leicester, UK
dr98@leicester.ac.uk

Jeremy Levesley
*School of Computing and Mathematical Sciences*
*University of Leicester*
Leicester, UK
jl1@leicester.ac.uk

Alexander N. Gorban
*School of Computing and Mathematical Sciences*
*University of Leicester*
Leicester, UK
ag153@leicester.ac.uk

Tim J. Coats
*Emergency Medicine Academic Group, Department of Cardiovascular Sciences*
*University of Leicester*
Leicester, UK
tc61@le.ac.uk

*Abstract*— Electronic patient records (EPRs) produce a wealth of data but contain significant missing information. Understanding and handling this missing data is an important part of clinical data analysis and if left unaddressed could result in bias in analysis and distortion in critical conclusions. Missing data may be linked to health care professional practice patterns and imputation of missing data can increase the validity of clinical decisions. This study focuses on statistical approaches for understanding and interpreting the missing data and machine learning based clinical data imputation using a single centre's paediatric emergency data and the data from UK's largest clinical audit for traumatic injury database (TARN). In the study of 56,961 data points related to initial vital signs and observations taken on children presenting to an Emergency Department, we have shown that missing data are likely to be non-random and how these are linked to health care professional practice patterns. We have then examined 79 TARN fields with missing values for 5,791 trauma cases. Singular Value Decomposition (SVD) and k-Nearest Neighbour (kNN) based missing data imputation methods are used and imputation results against the original dataset are compared and statistically tested. We have concluded that the 1NN imputer is the best imputation which indicates a usual pattern of clinical decision making: find the most similar patients and take their attributes as imputation.

## I. Introduction

The World Health Organisation (WHO) reported that as of June 2023, the mortality due to COVID-19 was over 6.9 million with over 768 million confirmed cases since the beginning of the pandemic [1]. This statistic comes from official cases that have been reported globally, however, it was also noted that the reported cases are not representative of inflection rates due to the reductions in testing and complete reported data or inconsistent certification of COVID-19 as cause of death [2]. It was reported in [3, pp. 130–137] that during 2020 and 2021 the excess mortality was estimated as 2.7 times greater than the official toll. Unfortunately, excess mortality is not collected or published by many countries due to the absence of electronic surveillance systems and death reporting in some locations. This did not mean that these areas were less severely affected. The data available only provided an underestimate of the true number of COVID cases and deaths in countries that required detailed reporting.

The data were needed to conduct analysis on pharmaceutical interventions and policies and strategies that governments had to develop to mitigate the situation in this uncertain time. With these extensive data gaps, mortality figures were suggested to be estimated relying on the data exists and other relevant factors collected from countries at regional and country level in the light of the missing data. In fact, a set of analytical methods are being developed to produce estimation of access deaths. In February 2021, Technical Advisory Group (TAG) was formed in collaboration with the WHO and the United Nations Department of Economic and Social Affair for mortality assessment [3, pp. 130–137]. The experts of TAG have studied the impact of the pandemic and how to quantify the impact using the existing surveys and consensus to fill in data gaps and in the absence of nationally representative data, methods that use subnational data were suggested for imputing the missing values.

The preceding example is an example of 'Dark Data': the data we don't have, they haven't been recorded, yet they have a major effect on critical decisions and actions [4]. Unless an awareness of dark data types and why/how the missing data arise and techniques to render some experiences visible, we may face disastrous global consequences.

Let's take another example of how missing data can have fatal effect in public health.

In April 2020, Centres for Disease Control and Prevention (CDC) team published a preliminary analysis of U.S. paediatric COVID cases [5]. Although the data from China suggested that paediatric cases might be less severe than cases in adults, serious characteristics among paediatric cases in U.S. could not be described due to absence of the data. The centre reported that at the time of the analysis, for the majority of paediatric cases (children age <18) in U.S., important characteristics such as disease symptoms, severity or underlying conditions were unknown due to high workload of health personnel involved in response activities. It was noted that the missing values for many variables are unlikely to be at random. Two crucial questions to be asked were: how did the clinical severity of COVID in children differ from adults and how the information could be extracted in the light of both known and unknown data. It was suggested that the results of the analysis must be cautiously interpreted as statistical comparison between age groups were not performed because of the high percentage of the missing information on important variables including hospitalisation status [5]. These information gaps made it much harder to marshal action for this age group.

The lack of information limited not only the generalisability of results but also the potential impact of treatment. Suppose that those paediatric cases with unknown characteristics are assumed to have same characteristics as adults. Any analysis of patients with known variables might be misleading relative to the overall population of paediatric COVID patients and thus the actions taken based on such analysis might be the wrong actions leading to incorrect prescriptions and treatment. Imagine the potential impact of incorrect treatment on the child mortality or development.

The pandemic has rethought us one important lesson: the data that we do not have can be crucial as the data we have in critical decision making. In fact, such situations arise quite often, especially, in clinical studies [4], [6, pp. 203–216]. The missing data arise in many different forms and there might be several reasons of missingness resulting from unavailability of the data or both human and/or machine errors. In the first place, the awareness of possibility that there might be something missing helps us avoid the risk of drawing incorrect conclusions. Second, examination of the types of missing data, patterns in hidden information and methods for filling the missing data can equip us to identify the problem and protect us from making poor decisions.

The recent explosion in the use of Electronic Patient Records (EPR) has opened new opportunities for research in improving data quality in healthcare studies. Data about patients and outcomes will be widely available and no longer the domain of analytics specialists. All Emergency Physicians will soon have access to large patient datasets on their desktops with powerful data analytical and modelling software packages. This has the potential to transform the way in which clinicians and managers can understand and change healthcare systems and practice. However, like the pitfalls that await the non-specialist when using a powerful statistical package, this new combination of big data, computer power and machine learning analytics/modelling packages also has the potential to give misleading results if critical information is ignored or missed [7, No. 3], [8].

'Big data' is often 'dirty data', and so medical data is. Even the largest clinical, trauma and care routine datasets contain large numbers of blank cells (missing data), but in the past the need for missing data expertise was confined to academic researchers. However as local EPR derived datasets become available all emergency physicians need a better understanding of missing data to critically appraise both local information and research papers. Simple methods of dealing with missing data such as complete case analysis (only analysing the cases with complete data) or simple imputation (using widely available software to 'fill in the blanks'), are easy to do with recent analytics packages. However, they are highly prone to introduce bias if used without the awareness of how to analyse missing data and how to minimise potential distortion of conclusions.

Emergency medicine has been in the forefront of training in critical appraisal and the use of evidence-based medicine. However, our skills in these areas are negated if the analysis that we have positively appraised is based on a dataset with a hidden bias due to missing data. Therefore, it is important to understand the types of missing data presented in EPRs and the methodologies to address the missing data. The aim of this paper is to discuss how to recognise missing data, identify its potential impact on the analysis results, suggest methods to compensate and methods to test the quality of data imputation.

In this paper, we start with examining unique aspects of missing data in EPRs. We then presented a case study of quantifying the missingness in children's vital signs. In [9, pp. 16–21], our analysis of missingness in paediatric practice has demonstrated how the missing data collected from e-observations are linked to healthcare professional practice patterns. The randomness of missing data was statistically tested. We identified the dependency between variables. It was concluded that in clinical practices missing data are likely to be non-random. This must be accounted in clinical data management systems with appropriate labelling of blank fields. Otherwise, the available data may not be representative of the cohort being studied and analytics (whether statistical, artificial intelligence or machine learning) may lead to bias in the analysis and incorrect conclusions.

Our findings trigged our curiosity to take a step further: investigation whether we can fill in or impute missing data using the existing variables with a certain degree of accuracy and include all information in decision making process. Therefore, our curiosity turns into research on missing data imputation for the largest trauma data set.

In the second part of this paper, we have analysed data from UK's largest clinical audit for traumatic injury database (TARN) and suggested methodologies for interpretation and imputation of missing data in medical practices. First, we have highlighted the various forms of missing data, methods to test randomness and data imputation methods. Second, we have examined 79 TARN fields with missing values for 5,791 trauma cases [18]. This is a common form of dark data: the data we know are missing, we know that patients have inputs, but we do not know what the values are. For 79 fields, we have applied SVD and kNN based missing data imputation methods and compared imputation results against the original dataset. The quality of the imputations is statistically tested using three statistical tests.

II. DATA SETS

*A. POPS data set*

The Paediatric Observation Priority Score (POPS) has been developed to assess the acuity of children presenting to urgent and emergency care environments [10, pp. 38–44]. It has been utilised in the Emergency Department at the Leicester Royal Infirmary since 2012. Both paper and digital forms are available. The total score (0-16) is generated as the combination of 8 physiological, behavioural and known-risk parameters: oxygen saturations (Sats), level of alertness (AVPU), extent of breathing difficulty (Breathing), background history (Other), nurse gut feeling (Gut Feeling), heart rate (Pulse), respiratory rate (RR) and temperature (Temp). A score 0, 1, or 2 is assigned to each response for variables, so the maximum possible total score is 16.

In our work of missing data analysis [9, pp. 16–21], 56,042 POPS e-observations with score variables were collected from a web-based application. Initial assessments of children's vital signs were completed by nurses in electronic format when children arrived at the emergency department. Not all POPS records were completed due to short-term computer errors or child being referred to immediate treatment without initial assessment.

*B. TARN data set*

The Trauma Audit and Research Network (TARN) database is UK's national clinical audit for traumatic injury and recognised as the largest trauma registry in Europe [11]. The network of hospitals which send information to TARN across the Europe is gradually increasing.

A total of 5,791 trauma cases with significant traumatic injuries were examined using data from the UK TARN registry for England and Wales from January 1, 2015 to 2020. We have analysed 79 TARN input fields with missing values [18] . The data includes records from patients of age between 0 and 101-year-old who sustain injury resulting in hospital admission for between 1 day to 365 days, including critical care admissions. Fields of identifiers and output fields were excluded from missing data analysis.

The database includes TARN core fields containing severity indicators for trauma cases (e.g. Injury Severity Score), rehabilitation prescription fields, questionnaire 1 fields (questionary completed at the date of discharge), questionnaire 2 fields (questionary completed 6 month after injury) and some calculated fields. In the database, data completeness in fields varies with a range between 0% to 43%.

### III. METHODOLOGY

In this section, we present types of missing data, methods for missing data imputation and methods to test the quality of imputation.

*A. Types of Missingness*

There are different patterns of missingness with different potentials for bias and different implications for the way in which the data can be analysed [12, pp. 1–37]. Missingness may be associated with either observed factors (things that are recorded in the dataset) or unobserved factors (things that are not recorded in the dataset). It is easier to find and compensate for patterns of missingness when they are related to observed factors. Missingness can be described as having one of the following patterns:

*1) Missing completely at random (MCAR):*

Here the missing data is completely independent of both observed and unobserved patient characteristics. In other words, the probability of missing is independent of the missing value (of this variable) and values of any other variables. An example is when staff sometimes just forget to record a patient's GCS. MCAR data has a low potential to bias the analysis as those with and without the GCS are otherwise similar.

*2) Missing at random (MAR):*

This category can be confusing because the word 'random' is used. The data is in fact missing for a reason – but that reason is not related to the data itself (for example, when the Emergency Department (ED) is very busy staff may not have time to enter a GCS into the EPR). In technical terms the property 'missing' is not related to the missing value (and so appears to be random when you look at the data field that you are interested in). The probability of missing is independent of the missing value (of this variable), but can depend on other variables. MAR data can be difficult to identify and can bias the analysis. However, if information has been collected about the factor that caused the data to be missing (for example if the 'busyness' of the ED was recorded in the GCS example) then this can be identified and adjusted for in the analysis.

*3) Missing not at randon (MNAR):*

In this case missing data is related to the missing value itself. In other words, the probability of missing depends on the missing value of this variable. For example, if a 'minors' patient is walking and talking ED staff may not record a GCS, as the patient is obviously well. The reason why the data was not recorded (cause of the missingness) is related to the missing variable itself (the GCS was normal). MNAR data is important to recognise as it causes bias and is difficult to adjust for in the analysis. For example, if all the high GCS patients are missing from a dataset, any imputation of GCS will cause bias, as the imputation algorithm has no data about high GCS patients.

*4) Obligatory absent data:*

For example, variable "Time to operation" must be absent for all patients who did not have an operation. This type of absent data is also called "Data which do not exist" or "Data which must be missing". This type is important as any attempts to impute these missing values will distort the input dataset and must not be done (for example imputing the pregnancy test results for men will make the dataset nonsense). In one sense this type of data is not missing (as it cannot exist) – but the blank cell created in the dataset presents the same issues for big data analytics as a blank cell due to missing data.

*5) Dark data:*

Dark data [4] are data you do not have. D. Hand introduced 15 types of dark data. One of the most interesting is type 2 data: this is data which is not known to be missing (for example the patients who have used the NHS data opt-out will not be present in the dataset). This type of missing data cannot be seen and cannot be assessed, so any impact cannot be known and no adjustment to the analysis can be made. The NHS data opt-out is about 5%, but varies widely across the country, giving the potential to create misleading results for both service analysis or research, without any indication that this has happened [13, No. 1]. This bias could occur even if the very best methods are used in the analysis within a publication that seems of the highest quality on critical appraisal.

The clinical dataset underlying an EPR can be thought of as a table with a row for each patient and a column for each variable in the EPR (this is an over-simplified, but useful description). There are many reasons why data might be missing in a routinely collected emergency department EPR dataset and depending on the reason the missing data can be classified into one of the above categories:

a. Patient too sick – staff cannot prioritise recording data. – Missing not at random

b. Patient too well – staff think recording data not relevant. – Missing not at random

c. Staff too busy – no time for data recording. – Missing at random if the variable doesn't relate to how busy

the ED is; however, missing variables related to how busy the department is (for example, time of triage or administration of a drug) may be subject to bias due to MNAR.

d. Data not available for staff to record – potential for bias depends on reason for non-availability e.g., low potential from random breakdown of near patient testing (MAR), but a higher potential (MAR) if there is a pattern to the machine breakdown (such as no technician at night).

e. Data not relevant to the patient – there are many thousands of potential tests and interventions within a medical dataset, but any one patient will only undergo a small subset. This means that the cells in the dataset relating to all of the non-performed test or procedures will be blank for that patient (anecdotally, hospital datasets have more than 90% missing data – because the vast majority of data fields are not relevant to a particular patient). These data points are absent rather than missing (as they were not generated) or obligatory missing data but as noted above, the blank cell in a dataset presents the same issues for data analytics as a blank cell due to other forms of missing data.

f. Staff not engaged - data recording requires additional work with no immediate benefit. For example, in the Emergency Care Data Set (ECDS) many clinicians simply code the main factor (such as main diagnosis or main comorbidity) rather than all of the details. This means that other information is missing in the dataset .

g. Patients not willing – some groups of patients may be less likely to communicate and provide information – a complex interaction of social, age related, societal, ethnic and gender-based factors.

h. Temporal change in data structure – patient datasets continuously evolve as changes in healthcare create changes in the data structure, such as the inclusion of a new test or other piece of information (for example, a frailty score is a relatively recent addition in the ECDS). This means that all cells for this variable in the database before the change will be blank. A good example of temporal change is a move from one EPR to another – which may have a different data structure. This is obligatory missing data. Again, it could be argued that this data is not missing as it was never recorded, but the same issues arise for big data analytics.

i. Withdrawn consent for data use - 5.4% of NHS patients in the UK have opted out of some uses of their data. This 'case deletion' not a random process and so may bias the remaining data; it would be considered MNAR and also Dark Data.

j. Deliberate manipulation of data – deletion of data through hacking or other malicious intent. This could be missing at random if the hack was completely non-discriminatory or missing not at random if the hacker specifically deleted certain information.

k. Data loss – in the complexity of healthcare data systems there is the potential for corruption or loss of data during processes such as transfer of data, backup, merging legacy systems or moving between EPRs. The data lost may relate to a specific time period or type of data and would then be considered missing not at random and therefore a potential cause of bias.

To understand the patterns of missingness, a missing data analysis needs to be performed before the data is analysed. This involves:

1. Quantifying the missingness for each variable. Also understand how missingness is distributed across the cases. For example, a dataset with 5% missing data could contain many cases with a little missing data or a few cases with a lot of missing data.

2. Developing an understanding of the meaning of the missing values. This may involve discussion with healthcare staff who understand the data collection process and data engineers who understand how the dataset was curated and extracted. These people often have knowledge that enables you to understand why some data is missing, for example a server failure on a particular day or the introduction of new data fields due to the change in the type of analyser used for near patient testing.

3. Understanding if some of the data should be missing, and if so, deciding how to handle this obligatory missingness in the analysis/modelling. For example, if pregnancy is an important risk factor any modelling (either statistical or artificial intelligence) will struggle with the missing data for male pregnancy tests. So, the problem can be avoided by understanding why this data is missing and deciding if some data engineering might be appropriate before analysis (such as creating a new field of 'Yes' or 'No' for pregnancy in all patients, which will not have missing data).

4. An exploratory analysis to look for patterns of non-random dependences within the missing data:

    a. For each variable divide the patients into those with missing and known data and tabulate the characteristics each group (all other variables). Uneven distribution of other variables between groups means that that there is more likely to be some systematic missingness (all other variables should be evenly distributed between the groups if data is missing at random).

    b. Test whether properties that are missing are dependent on other variables in the dataset. For example, to test the randomness of missing data in variable A in relation to variable B two groups of patients can be created depending on whether variable A is 'known' or 'missing'. If there is a random association between the missing data in variable A and variable B, the means and distribution of variable B will be the same in both groups. A t-test of means equality [14] can be used to make this comparison (the exact comparison depends on the proportion of missing data). A Mann-Whitney U test should be used if variable B is not normally distributed.

    However, it can still be difficult to find and compensate for patterns of missingness when they are related to observed factors – for example, if the relationship is highly complex (e.g., if the data is

missing due to an interaction of age, ethnicity, and social deprivation). There is also the danger that using this approach for very large datasets might detect statistically significant but very small differences between A and B, which are too small to impact on findings. Similarly small datasets may fail to detect differences between A and B which are meaningful.

   c. Evaluate whether missing data in one variable (A) is dependent on missing data in another variable (B). To do this, new binary variables 'MA' and 'MB' are created depending on the presence/absence of missing data for each variable and a chi squared test is used to assess the interdependence of missingness between the variables.

5. Evaluate missing outcome data. There is the potential to model outcomes [6, pp. 203–216] to substitute for missing outcomes, but handling missing outcome data is complex and can easily lead to error, for example, if all of the dead patients are missing from the dataset any attempt to model outcomes for these patients will be misleading. Modelling of missing outcomes needs specialist advice due to the high chance of introducing bias

*B. Machine Learning based missing data imputation*

In the literature, various methods have been proposed to handling the missing data including simple methods such as complete case analysis, ignoring and discarding the data and imputation. Methods like complete case analysis and ignoring and discarding the data have some drawbacks and should be carefully used as they can introduce bias.

Imputation methods aim to replace missing values with the estimated ones based on information available in datasets. The goal of the imputation is to obtain statistically valid inferences from the incomplete data. The methods identify the relationship in known values of the dataset to estimate the missing values of attribute of interest. The cost of imputation is usually less than the cost of collecting the data. There are various Imputation methods including *Case Substitution*, *Mean or Mode Imputation*, *Regression Imputation* and *Hot Deck and Cold Deck*.

In this work, we propose to employ k-Nearest Neighbour (kNN) and Singular Value Decomposition (SVD) based imputation to estimate and substitute missing data with imputed values.

*1) Nearest Neighbour Imputation*

k-nearest neighbour (kNN) based data imputation is one of widely used non-parametric imputation methods where missing value is replaced with a single estimated value or multiple plausible values that calculated from the k nearest observed data. The method is often referred as *nearest neighbour imputation* or *kNN imputation*.

In this work, we employ kNN imputation for single value imputation. The kNN imputation method selects patients' records with patients' profiles similar to record of interest to impute the missing values. If we consider attribute A has one missing value for a patient record, this method finds k other patients' records, which have a value present in other attributes. For our analysis, an average of values from k closest records is then used as an estimate for the missing value in attribute A. For the similarity measurement, Euclidean distance was used. We should emphasise that kNN imputation can be used even in case where there are no any complete records [17].

*2) SVD-based Imputation*

Singular Value Decomposition (SVD) based imputation is a non-parametric method which is free from any distributional or structural assumptions [15, pp. 31–39] [16, pp. 77–85]. SVD can be applied for data with missing values. In this case eigenvalues can be not orthogonal but can be used to calculate repaired data matrix. Values from repaired matrix can be used for missing values imputation. There is also iterative (or expectation -maximisation) method: create repaired data matrix, then combine this matrix with original one (take known values from original data and unknown from repaired one) and recalculate repaired matrix. Iterate until convergence (update difference is below the determined threshold).

*C. Testing the quality of imputation*

The next step after imputation of missing values is to compare the observed and imputed data, i.e., evaluate the quality of the imputation and test the statistical validity of procedures. We present three statistical tests to assess the difference between the original values in the dataset and the estimated values in the simulated incomplete dataset.

1. Two sample t-test can be performed on datasets to estimate the difference in mean with and without imputation. P-value is probability of observing by chance the same or greater absolute value of difference of mean values if both samples are sampled from populations with the same mean.

2. F-test compares two samples' variances and tests the significance of variance changes. P-value is probability of observing by chance the same or greater absolute value of difference of variances if both samples are sampled from populations with the same variance.

3. Two sample Kolmogorov-Smirnov (KS) test can be performed to compare the empirical distributions of the observed and imputed data. It is used to evaluate significance of difference of distributions of two samples. P-value is probability of observing by chance the same or greater KS statistics if both samples are sampled from populations with the same distribution.

IV. RESULTS

In medicine, the models to predict the population health outcomes are limited with an underlying effect of things that we cannot observe directly, i.e., *dark matter in medicine*. In this section, we focus on how quantifying the missing data can change the way the practitioners learn with the data that we do not have. We then propose ways to make 'invisible' to 'visible', replacing the missing data with estimated values.

*A. A case study of quantifying the missingness*

In this section, our attention is to demonstrate how the 'invisible' can be quantified and classified in order to use them to understand what missingness tells us about clinical practices. We focus analysing the children vital signs in ePOPS data and examining whether there were any patterns to the missing data [9, pp. 16–21].

TABLE I. MISSING DATA IN DATABASE FOR EACH FIELD (TOTALLY 56,042 RECORDS)

| Variable | | Missing Data | |
|---|---|---|---|
| | | Number | Fraction |
| Age | | 0 | 0.00% |
| POPS Variables | Breathing Score | 22,444 | 40.05% |
| | AVPU Score | 19,120 | 34.12% |
| | Gut Feeling Score | 19,944 | 35.59% |
| | Other Score | 20,060 | 35.79% |
| | Sats Score | 29,452 | 52.55% |
| | Pulse Score | 28,417 | 50.71% |
| | RR Score | 28,591 | 51.02% |
| | Temp Score | 28,540 | 50.93% |

TABLE II. EXPECTED AND OBSERVED NUMBER OF RECORDS WITH COMPLETE, PARTIALLY MISSING AND COMPLETELY MISSING POPS VARIABLES

| | Expected number of records | Observed number of records | Observed fraction of record | p-value |
|---|---|---|---|---|
| Complete | 67 | 25,114 | 45% | $< 10^{-300}$ |
| Partially missing | 55,461 | 12,707 | 23% | $< 10^{-300}$ |
| Completely missing | 514 | 18,221 | 32% | $< 10^{-300}$ |

First, we have tested the hypothesis that missed data in all variables are missed independently (observed at random). The fraction of missing data (Table 1) was used as the estimates of $p_i$: calculated the probabilities of observing (i) Complete (C) records (without any missing values in 8 variables), (ii) Completely Missing (CM) records (all individual score variables are missing), and (iii) Partially Missing records (partially missed data only (records with at least one known and at least one missed ePOPS variable). Results of calculations are presented in Table 2. We conclude that we must reject the hypothesis that data are missing independently (with p-value$< 10^{-300}$ from the chi-square test). So, it is unlikely that data are missing (observed) at random.

Second, we tested the hypothesis that the distribution of missing data for pairs of ePOPS variables are independent. Pearson correlation coefficients for each pair of POPS variables are presented in Table 3. We can conclude with 99% confidence that the intra-correlations (inside each of 2 groups: Breathing, AVPU, Gut feeling, Other and Sats, Pulse, RR, Temp) are greater than the inter-correlations (between groups). This means that the grouping of Breathing, AVPU, Gut Feeling and Other behaves in a different way than Sats, Pulse, RR and Temp in relation to the extent of missing data. The same pattern was also seen when repeating the procedure for only partially missing data only. From partially missing data analysis, we conclude that there are high correlations inside the group of Breathing, AVPU, Gut Feeling and Other (minimal PCC is 0.3) and inside the group of Sats, Pulse, RR, and Temp, exclude insignificant correlation of Temp with Sats.

TABLE IV. FRACTION OF MISSING VALUES IN SATS, PULSE, RR, AND TEMP WHICH CORRESPONDS TO NORMAL (0) VALUE OF BREATHING, AVPU, GUT FEELING AND OTHER AMONG ALL MISSED VALUES

| Normal Value of | Missing values of | | | | |
|---|---|---|---|---|---|
| | Sats | Pulse | RR | Temp | All |
| Breathing | 65% | 61% | 62% | 64% | 61% |
| AVPU | 94% | 94% | 94% | 96% | 96% |
| Gut Feeling | 84% | 83% | 83% | 84% | 85% |
| Other | 84% | 83% | 83% | 85% | 85% |
| All | 59% | 50% | 51% | 53% | 56% |

Finally, we analyse the relationship between a normal value (zero) from initial assessment and missing measured values (Table 4). The data show that if the initial assessment variables (Breathing, AVPU, Gut Feeling and Other) were normal (0) then there was a higher-than-expected chance that the measured variables (Sats, Pulse, RR and Temp) would be missing. This would fit with the clinical practice model that the assessment variables are much quicker to ascertain as they are slightly more subject and based on observation alone, whereas determination of measured variables requires additional work.

In our analysis of missingness in children's vital signs, we tested and demonstrated how the missing data are linked to health care professional practice patterns. We concluded that it is unlikely that data are missing at random. We identified the dependency of variables between one-another. Therefore, using only the available data may result in significant bias and misleading results as there is risk to avoid critical information.

Our findings led us to take a step further to investigate methodologies for missing data imputation and test how accurately the missing data can be estimated using the existing variables.

*B. Can we shed light on the dark?*

This section presents results of kNN and SVD based missing data imputation. Our aim with the missing data imputation is to preserve the information in the data and the relationship among variables while creating statistically valid inferences from the incomplete data. In other words, we aim to use all available data across the variables and produce estimates maintaining the natural variability in the estimated values having the same properties of the distributions as in original dataset.

TABLE III. PEARSON'S CORRELATION COEFFICIENT FOR EACH PAIR OF POPS VARIABLES, PINK BACKGROUNDS SHOW PAIRS WITH CORRELATION COEFFICIENT >0.8

| | Breathing | AVPU | Gut Feeling | Other | Sats | Pulse | RR | Temp |
|---|---|---|---|---|---|---|---|---|
| Breathing | -- | 0.87 | 0.90 | 0.89 | 0.76 | 0.78 | 0.78 | 0.76 |
| AVPU | 0.87 | -- | 0.96 | 0.94 | 0.67 | 0.69 | 0.69 | 0.67 |
| Gut Feeling | 0.90 | 0.96 | -- | 0.97 | 0.69 | 0.71 | 0.71 | 0.70 |
| Other | 0.89 | 0.94 | 0.97 | -- | 0.69 | 0.72 | 0.71 | 0.70 |
| Sats | 0.76 | 0.67 | 0.69 | 0.69 | -- | 0.96 | 0.95 | 0.90 |
| Pulse | 0.78 | 0.69 | 0.71 | 0.72 | 0.96 | -- | 0.98 | 0.93 |
| RR | 0.78 | 0.69 | 0.71 | 0.71 | 0.95 | 0.98 | -- | 0.93 |
| Temp | 0.76 | 0.67 | 0.70 | 0.70 | 0.90 | 0.93 | 0.93 | -- |

TABLE V. - *P*-VALUE OF T-TEST OF MEAN COMPARISON, F-TEST OF VARIANCE COMPARISON AND KOLMOGOROV-SMIRNOV
TEST OF DISTRIBUTION COMPARISON FOR OF SVD-BASED IMPUTATION, 1NN AND 3NN-BASED; GREEN BACKGROUND
HIGHLIGHT STATISTICALLY INSIGNIFICANTLY DIFFERENCES WITH SIGNIFICANCE LEVEL 99%

| Attribute | t-test of mean | | | F-test of variance | | | Kolmogorov-Smirnov test | | |
|---|---|---|---|---|---|---|---|---|---|
| | SVD | 1NN | 3NN | SVD | 1NN | 3NN | SVD | 1NN | 3NN |
| TC-06_GCS | 0.1401 | 0.2908 | 0.4731 | 0.8348 | 0.4143 | 0.0039 | <0.0001 | 0.9715 | 0.0668 |
| TC-08_LOScc | 0.9967 | 0.9835 | 0.9935 | 0.9958 | 0.9792 | 0.9988 | 1.0000 | 1.0000 | 1.0000 |
| TC-33_ED_GCS | <0.0001 | 0.2020 | 0.1465 | <0.0001 | <0.0001 | 0.6535 | <0.0001 | 0.9913 | 0.0004 |
| TC-34_ED_GCSEye | <0.0001 | 0.1286 | 0.1040 | <0.0001 | <0.0001 | 0.7798 | <0.0001 | 0.9997 | <0.0001 |
| TC-35_ED_GCSMotor | <0.0001 | 0.0239 | 0.0161 | <0.0001 | <0.0001 | 0.0026 | <0.0001 | 0.9997 | <0.0001 |
| TC-36_ED_GCSVerbal | <0.0001 | 0.6440 | 0.3298 | <0.0001 | 0.0085 | 0.1037 | <0.0001 | 1.0000 | <0.0001 |
| TC-37_ED_O2Sat | 0.4549 | 0.3497 | 0.4260 | 0.0296 | 0.0071 | <0.0001 | 0.1107 | 0.4447 | 0.0004 |
| TC-38_ED_Pulse | 0.0054 | 0.0491 | 0.0933 | 0.4788 | 0.1240 | 0.0001 | 0.0526 | 0.1228 | 0.0141 |
| TC-39_ED_RR | 0.6168 | 0.3228 | 0.1625 | 0.3417 | 0.3086 | <0.0001 | 0.0219 | 0.1420 | 0.0008 |
| TC-40_ED_SBP | 0.6755 | 0.8739 | 0.9824 | 0.0441 | 0.6179 | 0.0002 | 0.0964 | 0.9200 | 0.1684 |
| TC-41_PreHosp_GCS | 0.1683 | 0.1747 | 0.0869 | 0.0001 | 0.0006 | <0.0001 | <0.0001 | 0.1954 | <0.0001 |
| TC-42_PreHosp_GCSEye | <0.0001 | 0.0326 | 0.0212 | <0.0001 | <0.0001 | <0.0001 | <0.0001 | 0.9417 | 0.0080 |
| TC-43_PreHosp_GCSMotor | <0.0001 | 0.0029 | 0.0044 | <0.0001 | <0.0001 | <0.0001 | <0.0001 | 0.6876 | 0.0683 |
| TC-44_PreHosp_GCSVerbal | <0.0001 | 0.7070 | 0.1726 | 0.0073 | 0.0569 | <0.0001 | <0.0001 | 0.9298 | <0.0001 |
| TC-45_PreHosp_O2Sat | 0.5251 | 0.0238 | 0.0088 | <0.0001 | <0.0001 | <0.0001 | <0.0001 | 0.0406 | <0.0001 |
| TC-46_PreHosp_Pulse | 0.0031 | 0.3159 | 0.4651 | 0.0002 | 0.0002 | <0.0001 | 0.0042 | 0.0178 | <0.0001 |
| TC-47_PreHosp_RR | 0.0015 | 0.4706 | 0.2433 | <0.0001 | 0.0617 | <0.0001 | <0.0001 | 0.1028 | <0.0001 |
| TC-48_PreHosp_SBP | 0.5580 | 0.1837 | 0.2639 | <0.0001 | 0.0014 | <0.0001 | 0.0547 | 0.0231 | <0.0001 |
| NF-06 | <0.0001 | 0.2844 | 0.2486 | <0.0001 | 0.1385 | <0.0001 | <0.0001 | 0.9985 | 0.0014 |
| NF-07 | 0.3260 | 0.3857 | 0.4096 | 0.3144 | 0.6842 | 0.1939 | 0.1720 | 0.9972 | 0.1229 |
| NF=-08 | 0.0048 | <0.0001 | <0.0001 | 0.1077 | 0.0658 | <0.0001 | <0.0001 | <0.0001 | <0.0001 |
| NF-09 | 0.0360 | 0.0684 | 0.0721 | <0.0001 | 0.0671 | <0.0001 | <0.0001 | 0.6166 | <0.0001 |
| NF-10 | 0.9849 | 0.8465 | 0.8194 | 0.6727 | 0.7246 | 0.5511 | 0.8306 | 1.0000 | 1.0000 |
| NF-11 | 0.9464 | 0.3681 | 0.3608 | 0.0230 | 0.2968 | 0.0264 | <0.0001 | 0.9994 | 0.4182 |
| NF=-12 | 0.0819 | 0.0001 | 0.0001 | 0.0049 | 0.0023 | <0.0001 | <0.0001 | 0.0099 | <0.0001 |
| NF-18 | 0.9920 | 0.9191 | 0.9170 | 0.8217 | 0.9638 | 0.7654 | 1.0000 | 1.0000 | 1.0000 |
| NF-20 | <0.0001 | 0.3938 | 0.3480 | 0.0082 | 0.0036 | 0.4130 | <0.0001 | 1.0000 | 0.2699 |
| NF-21 | <0.0001 | 0.1834 | 0.1646 | 0.0077 | <0.0001 | 0.6324 | <0.0001 | 1.0000 | <0.0001 |
| NF-22 | <0.0001 | 0.6882 | 0.5321 | 0.0001 | 0.1037 | 0.0921 | <0.0001 | 1.0000 | <0.0001 |

The performance of imputation methods was assessed through three statistical tests: t-test to estimate the difference in mean with and without imputation, F-test to test the change in variances and Kolmogorov-Smirnov (KS) to evaluate significance of distribution differences of two samples.

For each of 79 variables, we have applied SVD-based imputation (function svdWithGaps.m in [17]) and kNN-based imputation (function kNNImpute.m in [17]) for k=1,3,5,7,9,11,13,15. Full tables with results are presented in [18]. The Table 5 presents part of the results of mean comparison of SVD-based and kNN-based imputations for SVD, 1NN and 3NN and for 29 attributes. It is necessary to emphasise that we are interested to have the same or at least insignificantly different distribution in imputed dataset. The insignificance of difference between original and imputed datasets increasing with increasing k (see supplemented tables [18]). This result is not unexpected and the best coincidence can be achieved for imputation by mean which completely destroyed variance and produce essentially different distribution.

Since the KS test compares the equality of distributions, the last three columns in Table 5 gives an idea about best imputer – the imputer which minimally change distribution. In this sense, the 1NN imputer is the best imputation. This result is also expected because it corresponds to usual practice of clinicians: find the most similar patients and take their attributes as imputation.

V. CONCLUSION AND DISCUSSION

Electronic Patient Records (EPR) have expended the availability of large data sets to that can be used to improve clinical practice, patients' outcomes and advance clinical decision making through research. The EPR data, however, often contain substantial missing information that creates challenges for interpretation and utilising the data for medical conclusions.

In this paper, we have addressed the problem of missing data that can have major effect on critical decisions in medicine. This work focused on demonstrating the importance of missing data analysis and proposing Machine Learning based approaches for missing data imputation based on two real world electronic patient records: POPS and TARN. Using these datasets, our study (1) improved the understanding of types of missing data in health care (2) increased the awareness of the challenges and how missingness can be interpreted in clinical practices, and (3) proposed methodologies for missing data imputation using non-missing information.

The POPS data were collected from e-observations completed by nurses on the initial assessment of children vital signs when they arrive at the Leicester Royal Infirmary children's emergency department. Our analysis showed that the pattern of missingness in POPS variables was not at random, i.e. there was a dependence of the distribution of missing data on the individual components of POPS. There were two groups of variables: heart rate, breathing rate, temperature and oxygen saturations forming one group and AVPU, Work of Breathing, Gut Feeling and Other forming a second grouping. Within each group having missing values were highly correlated. This grouping is linked to the usual clinical practice of staff taking observation. More subjective components are usually determined by looking at the patient during initial assessment. The second group of variables are needed to be measured and recorded later. Our data also showed that if the initial assessment variables are normal (0), then measured variables would be missing. This fits with the clinical practice model: undertaking full set of observation (vital sign measures) in real life is unlikely to occur for minor illnesses or injuries when staff feel that there is no need to record initial assessment as it is obvious that child is well.

Understanding of missing data, classification of missing data and how this can be linked to clinical practice patterns that improve patients' safety through more accurate immediate clinical decision making and managing large clinical data.

In the second part of this work, Trauma Audit and Research Network's (TARN) database was analysed. We applied Machine learning based data imputation methods and results were compared with the original dataset using statistical tests. We examined whether and how clinicians' perceptions may differ from ML – based decisions. Our findings showed that the 1NN imputer is the best imputation when compared with kNN for other k and SVD. This suggests that ML – guided decision making was congruent with the way clinicians typically made decisions in traumatic injuries, i.e., decisions made based on patterns being derived from the 'experience' learnt from patients having similar medical records.

In our study, we did not consider any controversy between heuristic decision making and machine generated data imputation. Instead, we approach ML-based data imputation from its link to heuristics involving pattern recognition where the patterns are derived from clinician's 'clinical experience'. Both essentially make decisions by finding the most similar overall patterns for the non-missing characteristics, so the imputation based on the same approach is likely to have a high 'face validity' for clinicians.